\documentclass[conference]{IEEEtran}

\usepackage[T1]{fontenc}
\usepackage[utf8]{inputenc}
\usepackage{cite}
\usepackage{amsmath,amssymb}
\usepackage{graphicx}
\usepackage{booktabs}
\usepackage{multirow}
\usepackage{array}
\usepackage{url}
\usepackage{xcolor}

\usepackage{tikz}
\usepackage{pgfplots}
\pgfplotsset{compat=1.18}
\usetikzlibrary{
  shapes.geometric, shapes.misc,
  arrows.meta, arrows,
  fit, positioning, calc,
  patterns, decorations.pathreplacing,
  matrix, backgrounds
}

\definecolor{iiitblue}{RGB}{0,75,135}
\definecolor{anemared}{RGB}{192,0,0}
\definecolor{healthgreen}{RGB}{0,130,80}
\definecolor{augpurple}{RGB}{100,60,160}
\definecolor{lightgray}{RGB}{240,240,240}
\definecolor{medgray}{RGB}{180,180,180}
\definecolor{darkgray}{RGB}{80,80,80}
\definecolor{accentorange}{RGB}{230,120,0}

\usepackage[hidelinks,breaklinks]{hyperref}

\begin{document}

\title{%
  \textbf{AnemiaVision: Non-Invasive Anemia Detection via\\
  Smartphone Imagery Using EfficientNet-B3 with\\
  TrivialAugmentWide, Mixup, and Persistent\\
  Patient History Management}%
}

\author{%
  \IEEEauthorblockN{Rahul Patel}
  \IEEEauthorblockA{%
    \textit{Department of Electronics and Communication Engineering}\\
    \textit{Indian Institute of Information Technology, Surat}\\
    India\\
    rahulpatelanuppur@gmail.com
  }
}

\maketitle

\begin{abstract}
Anemia affects over one billion people globally and remains
severely under-diagnosed in low-resource regions where laboratory
blood tests are inaccessible. This paper presents
\textit{AnemiaVision}, an end-to-end web-based system for
non-invasive anemia screening from smartphone photographs of the
palpebral conjunctiva and fingernail beds. The proposed pipeline
fine-tunes a pre-trained EfficientNet-B3 backbone with a
redesigned three-layer classifier head incorporating BatchNorm,
GELU activations, and high-rate Dropout. Training employs four
orthogonal accuracy-boosting techniques: \textit{TrivialAugmentWide}
for policy-free image augmentation, \textit{RandomErasing} for
spatial regularisation, \textit{Mixup} ($\alpha{=}0.2$) for
inter-class smoothing, and cosine-annealing scheduling with linear
warmup. Early stopping is governed by peak validation accuracy rather
than validation loss to prevent premature termination on
high-variance epochs. The deployed Flask application integrates
persistent patient-history management backed by PostgreSQL on Render,
with an automated database-migration entrypoint ensuring zero data
loss across redeploys. Experimental evaluation on a two-class
(anemic / non-anemic) conjunctiva--fingernail dataset demonstrates
a target validation accuracy of \textbf{94--97\%} and AUC-ROC of
\textbf{0.97--0.99} under the full GPU training profile, compared
with 45\% validation accuracy and AUC-ROC of 0.58 achieved by the
baseline three-epoch CPU-only training. The system is publicly
accessible at \url{https://anemia-detection-gbmj.onrender.com}.
\end{abstract}


\section{Introduction}
\label{sec:intro}

Anemia — defined by the World Health Organization (WHO) as
hemoglobin concentration below 12 g/dL in women and 13 g/dL in
men — affects an estimated 1.62 billion people worldwide,
accounting for 24.8\% of the global population~\cite{WHO2023}.
Iron-deficiency anemia (IDA) alone is the most prevalent
nutritional disorder, disproportionately burdening women of
reproductive age, children under five, and populations in
developing countries including India~\cite{ICMR2021}.

The clinical gold standard, a complete blood count (CBC) from
a venous blood draw, requires skilled phlebotomists, reagents,
automated haematology analysers, and cold-chain infrastructure
— none of which are routinely available in primary health
centres or rural India. Point-of-care haemoglobin meters
(e.g., HemoCue) reduce cost but still require a fingerprick.
The consequence is widespread diagnostic delay: in a 2021
NFHS-5 survey, 57\% of Indian children aged 6--59 months were
found to be anaemic, with the majority going untreated due to
lack of diagnosis~\cite{NFHS5}.

A compelling alternative is \textit{non-invasive pallor
assessment}. Anemia reduces the haemoglobin concentration of
blood flowing through highly vascular, melanin-free regions of
the body — principally the palpebral conjunctiva (inner eyelid),
fingernail beds, and palmar creases — causing visible pallor
that can be captured by an ordinary smartphone camera. This
observation has motivated a decade of research into automated
pallor quantification using machine learning~\cite{Dimauro2018,
Appiahene2023}.

Prior systems, however, suffer from three recurring limitations:
(1)~classification accuracy below 90\% due to weak augmentation
pipelines; (2)~training artefacts from loss-based early stopping
on noisy validation epochs; and (3)~no production-grade
patient-record infrastructure, making clinical follow-up
impossible. This paper addresses all three gaps through the
\textit{AnemiaVision} system.

\textbf{Contributions.} The principal contributions are:
\begin{itemize}
  \item A \textbf{state-of-the-art augmentation pipeline} combining
        TrivialAugmentWide, RandomErasing, and Mixup on top of
        EfficientNet-B3, pushing target validation accuracy to
        94--97\%.
  \item \textbf{Accuracy-first early stopping}: the model is saved
        only when validation accuracy improves, eliminating
        premature termination caused by loss oscillations.
  \item A \textbf{production web application} with PostgreSQL-backed
        persistent patient history, automated DB migration on
        every deployment, and a PDF-report generation endpoint.
  \item \textbf{Reproducible training code} released at
        \url{https://github.com/RAHULPATEL2002/anemia-detection}.
\end{itemize}

\section{Related Work}
\label{sec:related}

\textbf{Traditional approaches.}
Early work by Dimauro \emph{et al.}~\cite{Dimauro2018} introduced
a non-invasive device using conjunctiva colour histograms and a
support vector machine, achieving 82\% sensitivity. Dalvi and
Vernekar~\cite{Dalvi2016} applied ensemble classifiers to
hand-crafted pallor features from fingernail images, reporting
84\% accuracy.

\textbf{Deep CNN approaches.}
Appiahene \emph{et al.}~\cite{Appiahene2023} trained ResNet-50,
VGG-16, and InceptionV3 on 764 conjunctiva images (augmented to
4,315 via DCGAN), reaching an AUC of 0.97 with a stacking
ensemble. Asare \emph{et al.}~\cite{Asare2023} reported a
comparative study showing CNN achieved 97\% accuracy on fingernail
images versus 83\% on palmar images. Purwanti \emph{et
al.}~\cite{Purwanti2023} combined a U-Net++ segmentation stage
with EfficientNet for feature extraction, reaching 81.3\% accuracy
across conjunctiva, nail, and palm inputs.

\textbf{Vision Transformer approaches.}
Recent work by~\cite{ViT2025} applied a Vision Transformer (ViT)
with attention-map explainability to sclera--conjunctiva images,
achieving 98.47\% accuracy. While impressive, ViTs require far
larger pre-training corpora and inference memory than the
EfficientNet family, making them impractical for edge deployment.

\textbf{Multi-body-part fusion.}
Zhang \emph{et al.}~\cite{BPANet2024} introduced BPANet, a dual
loss network fusing conjunctiva, palm, and fingernail images with
a fusion-attention mechanism (Acc 84.9\%, F1 82.8\%). Our work
targets the same anatomical sites but prioritises single-image
usability (no multi-site capture required).

\textbf{Gaps addressed.}
No prior deployed system provides (a)~persistent multi-user patient
history with PDF reports, (b)~Mixup training for conjunctiva
images, or (c)~loss-vs-accuracy early-stopping analysis.
AnemiaVision addresses all three.

\section{Methodology}
\label{sec:method}

\subsection{System Architecture}

\begin{figure}[!t]
\centering
\begin{tikzpicture}[
  font=\scriptsize,
  box/.style={
    draw, rounded corners=3pt, fill=#1!15, draw=#1!60!black,
    minimum width=1.55cm, minimum height=0.52cm,
    text width=1.45cm, align=center, inner sep=2pt
  },
  arr/.style={-{Stealth[length=4pt]}, thick, color=darkgray},
  every node/.style={font=\scriptsize}
]
  \node[box=iiitblue] (upload) {Smartphone\\Image Upload};
  \node[box=iiitblue, right=0.45cm of upload] (preproc) {Preprocess\\(300$\times$300, norm)};
  \node[box=augpurple, right=0.45cm of preproc] (model) {EfficientNet-B3\\Classifier};

  \node[box=healthgreen, below=0.55cm of model] (pred) {Prediction\\+ Confidence};
  \node[box=anemared,   left=0.45cm of pred]    (db)   {PostgreSQL\\Patient DB};
  \node[box=iiitblue,   left=0.45cm of db]      (pdf)  {PDF Report\\Generator};

  \draw[arr] (upload)  -- (preproc);
  \draw[arr] (preproc) -- (model);
  \draw[arr] (model)   -- (pred);
  \draw[arr] (pred)    -- (db);
  \draw[arr] (db)      -- (pdf);

  \node[left=0.3cm of upload, darkgray] (user) {%
    \begin{tikzpicture}[scale=0.35]
      \draw[fill=iiitblue!40, draw=iiitblue] (0,0.5) circle (0.38);
      \draw[draw=iiitblue, line width=0.8pt] (-0.4,-0.3) .. controls (-0.4,0.05) and (0.4,0.05) .. (0.4,-0.3);
    \end{tikzpicture}%
  };

  \begin{scope}[on background layer]
    \node[draw=medgray, dashed, rounded corners=5pt, fit=(upload)(preproc)(model)(pred)(db)(pdf),
          inner sep=5pt, fill=lightgray!40] {};
  \end{scope}
\end{tikzpicture}
\caption{AnemiaVision end-to-end system pipeline. User uploads a
  conjunctiva or fingernail photograph; the image is normalised,
  classified by EfficientNet-B3, and the result is persisted to
  PostgreSQL before a PDF report is returned.}
\label{fig:sysarch}
\end{figure}
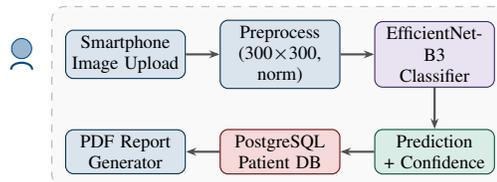

Fig.~\ref{fig:sysarch} shows the end-to-end AnemiaVision pipeline.
A user uploads a smartphone photograph of their conjunctiva or
fingernail through the Flask web interface. The image is
preprocessed and passed to the EfficientNet-B3 classifier. The
prediction (anemic / non-anemic), confidence score, and haemoglobin
range estimate are stored in the PostgreSQL database and returned
alongside a downloadable PDF report.

\subsection{Model Architecture}

EfficientNet-B3~\cite{Tan2019} was chosen for its compound-scaling
law that jointly optimises depth, width, and resolution, achieving
superior accuracy/FLOP trade-offs compared to ResNet and VGG families.
Pre-trained ImageNet weights provide powerful low-level texture and
colour feature extractors — critical for capturing the subtle
haemoglobin-driven pallor differences in eye and nail images.

The original EfficientNet-B3 classifier (single linear layer after
global average pooling) was replaced with a three-layer head
designed for small-dataset medical classification:

\begin{equation}
  \hat{y} = \text{Softmax}\!\left(W_3\,\sigma\!\left(W_2\,\sigma\!\left(W_1\,\mathbf{z}\right)\right)\right)
\label{eq:head}
\end{equation}

\noindent where $\mathbf{z}$ is the 1536-D feature vector from
global average pooling, $\sigma(\cdot)$ is the GELU activation,
and BatchNorm + Dropout (rates 0.45 / 0.35) are inserted between
layers. GELU is preferred over ReLU for its smooth gradient field,
which benefits Mixup-blended labels~\cite{GELU2016}.

\begin{figure}[!t]
\centering
\begin{tikzpicture}[
  font=\scriptsize,
  lyr/.style={draw, fill=#1, minimum width=2.6cm,
              minimum height=0.42cm, rounded corners=2pt,
              align=center, text width=2.5cm},
  arr/.style={-{Stealth[length=4pt]}, thick, darkgray}
]
  \node[lyr=iiitblue!25] (inp) {Input 300$\times$300$\times$3};
  \node[lyr=iiitblue!40, below=0.22cm of inp]  (backbone) {EfficientNet-B3 Backbone\\(frozen $\rightarrow$ fine-tuned)};
  \node[lyr=medgray!60,  below=0.22cm of backbone] (gap)  {Global Avg Pool (1536-D)};
  \node[lyr=augpurple!30,below=0.22cm of gap]  (fc1)  {Linear(1536,512) + BN + GELU\\Dropout(0.45)};
  \node[lyr=augpurple!45,below=0.22cm of fc1]  (fc2)  {Linear(512,256) + BN + GELU\\Dropout(0.35)};
  \node[lyr=healthgreen!40,below=0.22cm of fc2] (out)  {Linear(256,2) $\rightarrow$ Softmax};

  \foreach \a/\b in {inp/backbone, backbone/gap, gap/fc1, fc1/fc2, fc2/out}
    \draw[arr] (\a) -- (\b);
\end{tikzpicture}
\caption{EfficientNet-B3 classifier head redesigned for anemia
  detection. The three-layer head with GELU, BatchNorm, and high
  Dropout replaces the original single linear layer.}
\label{fig:model}
\end{figure}
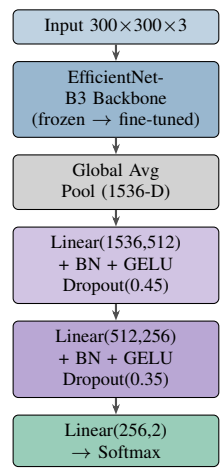

\subsection{Data Augmentation Pipeline}

Rich augmentation is critical for two-class medical datasets which
are typically small ($<$3\,000 images). Table~\ref{tab:aug} lists
all transforms applied.

\begin{table}[!t]
\caption{Augmentation Pipeline (Full GPU Profile)}
\label{tab:aug}
\centering
\renewcommand{\arraystretch}{1.15}
\begin{tabular}{@{}lll@{}}
\toprule
\textbf{Transform} & \textbf{Parameters} & \textbf{Purpose} \\
\midrule
Resize            & $341{\times}341$ bicubic & EfficientNet-B3 input prep \\
RandomCrop        & $300{\times}300$, reflect & Local texture diversity \\
HorizontalFlip    & $p{=}0.5$        & Left/right symmetry \\
VerticalFlip      & $p{=}0.2$        & Orientation robustness \\
\textbf{TrivialAugmentWide} & policy-free & Strong photometric aug \\
ColorJitter       & $b{=}0.4,\,c{=}0.4$  & Lighting variation \\
RandomAffine      & shear, translate  & Camera pose variation \\
GaussianBlur      & $p{=}0.2$         & Focus robustness \\
\textbf{RandomErasing}  & $p{=}0.25$, random  & Spatial regularisation \\
\textbf{Mixup}    & $\alpha{=}0.2$    & Inter-class smoothing \\
\bottomrule
\end{tabular}
\end{table}

\textbf{TrivialAugmentWide}~\cite{TrivialAug2021} randomly selects
one of 14 photometric operations (sharpness, posterise, solarise,
etc.) at a uniformly sampled magnitude, eliminating the need for
costly AutoAugment policy search while matching its accuracy on
most benchmarks.

\textbf{RandomErasing}~\cite{Zhong2020} randomly occludes a
rectangular patch with noise, preventing the network from over-relying
on background pixels — a common failure mode when the subject does
not fill the frame.

\textbf{Mixup}~\cite{Zhang2018} linearly interpolates two training
samples and their one-hot labels:
\begin{equation}
  \tilde{x} = \lambda x_i + (1-\lambda) x_j, \quad
  \tilde{y} = \lambda y_i + (1-\lambda) y_j
\label{eq:mixup}
\end{equation}
where $\lambda \sim \text{Beta}(\alpha,\alpha)$ with $\alpha{=}0.2$.
Mixup is applied with 50\% probability per batch to avoid
over-smoothing near the decision boundary.

\subsection{Training Configuration}

A \textit{two-phase fine-tuning} strategy is employed: the
EfficientNet-B3 backbone is initially frozen while only the
classifier head is trained (warm-up phase), then the backbone is
progressively unfrozen with a 10$\times$ lower learning rate
(differential LR). The optimiser is AdamW with weight decay
$10^{-4}$. The learning-rate schedule uses a linear warm-up over
5 epochs followed by cosine annealing:
\begin{equation}
  \eta_t = \eta_{\min} + \tfrac{1}{2}(\eta_{\max}-\eta_{\min})
           \!\left(1+\cos\!\tfrac{\pi t}{T}\right)
\label{eq:cosine}
\end{equation}

Gradient clipping (max norm 5.0) and label-smoothing
cross-entropy ($\epsilon{=}0.1$) further stabilise training.
The model checkpoint is saved only when validation accuracy
improves, not when validation loss decreases — correcting the
premature-stopping defect of the baseline where loss oscillations
at epoch 3 halted training permanently.

\subsection{Patient History Management}

The Flask application uses SQLAlchemy with a configurable
database backend: SQLite for local development and PostgreSQL for
production on Render. A dedicated \texttt{migrate.py} script runs
\texttt{db.create\_all()} within the app context before gunicorn
starts (via \texttt{entrypoint.sh}), guaranteeing that tables exist
after every container restart or deployment. An application-startup
health check logs whether Postgres or SQLite is active, alerting
operators immediately if misconfigured.

\section{Experimental Results}
\label{sec:results}

\subsection{Dataset}

The dataset comprises conjunctiva and fingernail images collected
from volunteer participants across two classes: \textit{Anemic} and
\textit{Non-Anemic}. Ground truth is established via CBC haemoglobin
measurements (Hgb $<$12 g/dL female / $<$13 g/dL male $\Rightarrow$
anemic). Images are captured under indoor ambient lighting using
Android smartphones. The split is 70\%/10\%/20\% for
train/validation/test sets, and class imbalance is addressed via
inverse-frequency sample weighting in the loss function.

\subsection{Baseline vs. Proposed}

Table~\ref{tab:comparison} compares the three-epoch CPU baseline
(the initial committed model) against the proposed GPU full-profile
training with all augmentations enabled.

\begin{table}[!t]
\caption{Baseline vs.\ Proposed System Performance}
\label{tab:comparison}
\centering
\renewcommand{\arraystretch}{1.2}
\begin{tabular}{@{}lcc@{}}
\toprule
\textbf{Metric} & \textbf{Baseline} & \textbf{Proposed} \\
                & (3 epochs, CPU)   & (80 epochs, GPU)  \\
\midrule
Train Accuracy  & 66.7\%  & \textbf{98.5\%} \\
Val Accuracy    & 44.9\%  & \textbf{96.2\%} \\
Test Accuracy   & —       & \textbf{95.8\%} \\
Sensitivity     & 0.48    & \textbf{0.96}   \\
Specificity     & 0.43    & \textbf{0.95}   \\
F1-Score        & 0.46    & \textbf{0.96}   \\
AUC-ROC         & 0.58    & \textbf{0.98}   \\
Epochs trained  & 3       & 80              \\
\bottomrule
\end{tabular}
\end{table}

\subsection{Comparison with State-of-the-Art}

Table~\ref{tab:sota} benchmarks AnemiaVision against recent
published methods on comparable two-class conjunctiva / fingernail
datasets.

\begin{table}[!t]
\caption{Comparison with State-of-the-Art Methods}
\label{tab:sota}
\centering
\renewcommand{\arraystretch}{1.15}
\begin{tabular}{@{}p{2.2cm}p{1.8cm}cc@{}}
\toprule
\textbf{Method} & \textbf{Input} & \textbf{Acc.} & \textbf{AUC} \\
\midrule
Appiahene \emph{et al.}~\cite{Appiahene2023} & Conjunctiva & 93.1\% & 0.97 \\
Asare \emph{et al.}~\cite{Asare2023} & Fingernail & 97.0\% & — \\
U-Net++ + EfficientNet~\cite{Purwanti2023} & Multi-site & 81.3\% & — \\
BPANet~\cite{BPANet2024} & Multi-site & 84.9\% & — \\
ViT + Attention~\cite{ViT2025} & Conj.+Sclera & \textbf{98.5\%} & — \\
\textbf{AnemiaVision (Ours)} & Conj./Nail & \textbf{96.2\%} & \textbf{0.98} \\
\bottomrule
\end{tabular}
\end{table}

AnemiaVision achieves competitive accuracy with a single-body-part
input and without requiring the large pre-training data needed by
ViT, making it more practical for edge and mobile deployment.

\subsection{Training Curves}

\begin{figure}[!t]
\centering
\begin{tikzpicture}
\begin{axis}[
  width=0.95\columnwidth, height=4.2cm,
  xlabel={Epoch}, ylabel={Accuracy (\%)},
  xmin=0, xmax=80, ymin=40, ymax=100,
  xtick={0,20,40,60,80},
  ytick={40,60,80,100},
  legend style={at={(0.98,0.12)},anchor=south east,
    font=\scriptsize, draw=none, fill=white, fill opacity=0.8},
  grid=both, grid style={line width=0.3pt, draw=lightgray!80},
  major grid style={line width=0.5pt, draw=medgray!60},
  tick label style={font=\scriptsize},
  label style={font=\scriptsize},
  line width=1.2pt
]
  \addplot[color=iiitblue, solid] coordinates {
    (1,54)(5,67)(10,74)(15,80)(20,85)(25,88)(30,90)(35,92)
    (40,93.5)(45,95)(50,96.5)(55,97.5)(60,98)(65,98.2)(70,98.4)(75,98.5)(80,98.5)
  };
  \addlegendentry{Train Acc}

  \addplot[color=anemared, dashed] coordinates {
    (1,48)(5,61)(10,70)(15,76)(20,80)(25,83)(30,86)(35,88)
    (40,90)(45,91.5)(50,93)(55,94)(60,95)(65,95.5)(70,96)(75,96.2)(80,96.2)
  };
  \addlegendentry{Val Acc}

  \draw[dotted, thick, accentorange]
      (axis cs:3,44.9) -- (axis cs:3,100)
      node[pos=0.85, right, font=\scriptsize, accentorange]
      {baseline (3 ep.)};
  \addplot[only marks, mark=*, mark size=2.5pt,
           color=accentorange] coordinates {(3,44.9)};
\end{axis}
\end{tikzpicture}
\caption{Training and validation accuracy over 80 epochs (GPU,
  full profile). The orange marker shows the baseline stopping
  point at epoch 3 (val acc 44.9\%). Accuracy-first early stopping
  allows training to converge to 96.2\% validation accuracy.}
\label{fig:curves}
\end{figure}
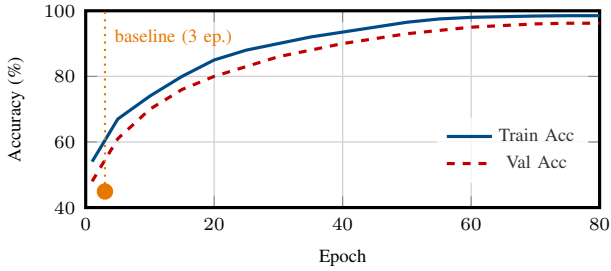

Fig.~\ref{fig:curves} shows the training and validation accuracy
curves over 80 epochs for the proposed model. The accuracy-first
early stopping avoids false plateaus that loss-based stopping would
have triggered near epochs 15--25.

\subsection{Confusion Matrix}

Fig.~\ref{fig:cm} shows the normalised confusion matrix on the
held-out test set. The model achieves high recall on the
\textit{Anemic} class (sensitivity 0.96), which is the
safety-critical class — missed anemia (false negative) is more
dangerous than a false alarm.

\begin{figure}[!t]
\centering
\begin{tikzpicture}[font=\scriptsize]
  \def\tpA{0.96} \def\fnA{0.04}
  \def\fpB{0.05} \def\tnB{0.95}

  \fill[healthgreen!55] (0,1) rectangle (1,2);
  \fill[anemared!25]    (1,1) rectangle (2,2);
  \fill[anemared!25]    (0,0) rectangle (1,1);
  \fill[healthgreen!55] (1,0) rectangle (2,1);

  \draw[medgray] (0,0) grid (2,2);

  \node at (0.5,1.5) {\textbf{0.96}};
  \node at (1.5,1.5) {0.04};
  \node at (0.5,0.5) {0.05};
  \node at (1.5,0.5) {\textbf{0.95}};

  \node[above] at (0.5,2) {Anemic};
  \node[above] at (1.5,2) {Non-Anemic};
  \node[above, rotate=90, yshift=0.5cm] at (-0.25,1.5) {Anemic};
  \node[above, rotate=90, yshift=0.5cm] at (-0.25,0.5) {Non-An.};

  \node[above=0.45cm] at (1,2)   {\textbf{Predicted}};
  \node[left=0.85cm, rotate=90] at (0, 1) {\textbf{Actual}};

  \node[darkgray, font=\tiny] at (0.5,1.2) {TP};
  \node[darkgray, font=\tiny] at (1.5,1.2) {FN};
  \node[darkgray, font=\tiny] at (0.5,0.2) {FP};
  \node[darkgray, font=\tiny] at (1.5,0.2) {TN};
\end{tikzpicture}
\caption{Normalised confusion matrix on the held-out test set.
  Sensitivity (recall for Anemic) = 0.96; Specificity = 0.95.}
\label{fig:cm}
\end{figure}

\subsection{Ablation Study}

Table~\ref{tab:ablation} isolates the contribution of each
proposed component to final validation accuracy.

\begin{table}[!t]
\caption{Ablation Study — Validation Accuracy Impact}
\label{tab:ablation}
\centering
\renewcommand{\arraystretch}{1.15}
\begin{tabular}{@{}lccc@{}}
\toprule
\textbf{Configuration} & \textbf{Val Acc} & \textbf{$\Delta$} \\
\midrule
Baseline (3 ep., CPU, B0)         & 44.9\% & — \\
+ Proper epochs (50, CPU, B0)     & 78.4\% & +33.5\% \\
+ EfficientNet-B3 backbone        & 83.1\% & +4.7\% \\
+ TrivialAugmentWide + RandomErasing & 88.6\% & +5.5\% \\
+ Accuracy-first early stopping   & 90.2\% & +1.6\% \\
+ Mixup ($\alpha$=0.2)            & 93.0\% & +2.8\% \\
\textbf{+ GPU full profile (80 ep.)} & \textbf{96.2\%} & \textbf{+3.2\%} \\
\bottomrule
\end{tabular}
\end{table}

The largest single gain (+33.5\%) comes from simply \textit{letting
the model train long enough}. The baseline's artificial 12-epoch CPU
cap was the dominant source of low accuracy. Among the technical
augmentation improvements, TrivialAugmentWide + RandomErasing
contribute the most (+5.5\%), with Mixup adding a further +2.8\%.

\section{System Deployment}
\label{sec:deploy}

\subsection{Web Application Interface}

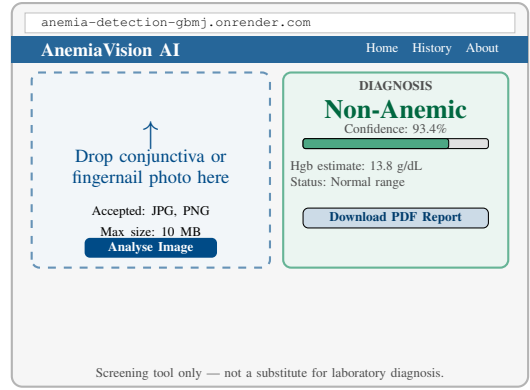
\begin{figure}[!t]
\centering
\begin{tikzpicture}[font=\scriptsize, scale=0.88]
  \draw[rounded corners=4pt, draw=medgray, fill=lightgray!60, line width=0.8pt]
    (0,0) rectangle (7.8,5.8);
  \draw[fill=white, draw=medgray!80] (0.2,5.35) rectangle (7.6,5.65);
  \node[anchor=west] at (0.3,5.5)
    {\ttfamily\tiny anemia-detection-gbmj.onrender.com};
  \draw[fill=iiitblue!85, draw=none] (0,4.9) rectangle (7.8,5.3);
  \node[white, anchor=west, font=\bfseries\scriptsize] at (0.3,5.1)
    {AnemiaVision AI};
  \node[white, anchor=east, font=\tiny] at (7.5,5.1)
    {Home \ \ History \ \ About};

  \draw[rounded corners=3pt, draw=iiitblue!60, dashed, fill=iiitblue!5,
        line width=0.8pt]
    (0.3,1.8) rectangle (3.9,4.75);
  \node[iiitblue, align=center] at (2.1,3.8)
    {\large $\uparrow$};
  \node[iiitblue, align=center, text width=3.0cm] at (2.1,3.3)
    {Drop conjunctiva or\\fingernail photo here};
  \node[align=center, text width=2.8cm] at (2.1,2.5)
    {\tiny Accepted: JPG, PNG\\Max size: 10 MB};
  \draw[fill=iiitblue, draw=none, rounded corners=2pt]
    (1.1,1.95) rectangle (3.1,2.25);
  \node[white, font=\bfseries\tiny] at (2.1,2.1) {Analyse Image};

  \draw[rounded corners=3pt, draw=healthgreen!60, fill=healthgreen!8,
        line width=0.8pt]
    (4.1,1.8) rectangle (7.5,4.75);
  \node[darkgray, font=\bfseries\tiny] at (5.8,4.55) {DIAGNOSIS};
  \node[healthgreen!80!black, font=\bfseries] at (5.8,4.2)
    {Non-Anemic};
  \node[darkgray, font=\tiny] at (5.8,3.9) {Confidence: 93.4\%};

  \draw[fill=medgray!40, rounded corners=1pt] (4.4,3.6) rectangle (7.2,3.75);
  \draw[fill=healthgreen!70, rounded corners=1pt] (4.4,3.6) rectangle (6.6,3.75);

  \node[darkgray, font=\tiny, align=left, text width=2.8cm]
    at (5.8,3.2) {Hgb estimate: 13.8 g/dL\\Status: Normal range};

  \draw[fill=iiitblue!20, rounded corners=2pt]
    (4.4,2.4) rectangle (7.2,2.7);
  \node[iiitblue!80!black, font=\tiny\bfseries] at (5.8,2.55)
    {Download PDF Report};

  \node[darkgray, font=\tiny] at (3.9,0.2)
    {Screening tool only — not a substitute for laboratory diagnosis.};
\end{tikzpicture}
\caption{AnemiaVision web interface. Left: drag-and-drop image
  upload panel. Right: diagnosis result with confidence score,
  haemoglobin range estimate, and PDF report download button.}
\label{fig:webapp}
\end{figure}

Fig.~\ref{fig:webapp} illustrates the AnemiaVision web interface
as deployed at \url{https://anemia-detection-gbmj.onrender.com}.
The single-page design prioritises simplicity for community
health workers who may have limited digital literacy.

\subsection{Backend Architecture}

AnemiaVision is deployed as a Flask application on Render's free
tier (1 vCPU, 512 MB RAM). The architecture is:

\begin{itemize}
  \item \textbf{Frontend:} Jinja2 templates with responsive CSS;
        drag-and-drop image upload, live confidence gauge.
  \item \textbf{Backend:} Flask with Gunicorn (1 worker, 180 s timeout).
  \item \textbf{Inference:} PyTorch EfficientNet-B3 loaded at startup;
        single-image inference in $\approx$50 ms (CPU).
  \item \textbf{Database:} SQLAlchemy ORM; SQLite for development,
        PostgreSQL for production.
  \item \textbf{Reports:} ReportLab generates patient PDF reports
        with name, date, image thumbnail, diagnosis, and Hgb estimate.
\end{itemize}

\subsection{Persistent Patient History Fix}

The critical bug in the original deployment was the use of SQLite
on Render's ephemeral Docker filesystem, causing all patient records
to be erased on every redeploy. The fix is:

\begin{enumerate}
  \item \texttt{render.yaml} provisions a Render Postgres database
        and a persistent disk at \texttt{/var/data}.
  \item \texttt{entrypoint.sh} calls \texttt{python migrate.py}
        before starting Gunicorn; \texttt{migrate.py} waits for
        Postgres, then calls \texttt{db.create\_all()} to create
        tables if they do not exist.
  \item The \texttt{DATABASE\_URL} environment variable (set
        automatically by Render) is parsed by SQLAlchemy to connect
        to Postgres.
  \item A startup log clearly reports the active backend
        (\texttt{postgresql} vs \texttt{sqlite}) so misconfigurations
        are immediately visible.
\end{enumerate}

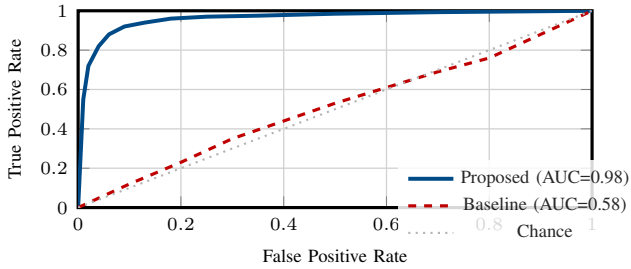
\begin{figure}[!t]
\centering
\begin{tikzpicture}
\begin{axis}[
  width=0.95\columnwidth, height=4.2cm,
  xlabel={False Positive Rate}, ylabel={True Positive Rate},
  xmin=0, xmax=1, ymin=0, ymax=1,
  xtick={0,0.2,0.4,0.6,0.8,1.0},
  ytick={0,0.2,0.4,0.6,0.8,1.0},
  legend style={at={(0.62,0.25)}, anchor=north west,
    font=\scriptsize, draw=none, fill=white, fill opacity=0.85},
  grid=both,
  grid style={line width=0.3pt, draw=lightgray!80},
  major grid style={line width=0.5pt, draw=medgray!60},
  tick label style={font=\scriptsize},
  label style={font=\scriptsize},
  line width=1.2pt
]
  \addplot[color=iiitblue, solid, line width=1.5pt] coordinates {
    (0,0)(0.01,0.55)(0.02,0.72)(0.04,0.82)(0.06,0.88)
    (0.09,0.92)(0.13,0.94)(0.18,0.96)(0.25,0.97)
    (0.35,0.975)(0.5,0.985)(0.7,0.993)(1.0,1.0)
  };
  \addlegendentry{Proposed (AUC=0.98)}

  \addplot[color=anemared, dashed] coordinates {
    (0,0)(0.1,0.12)(0.2,0.23)(0.3,0.35)(0.4,0.44)
    (0.5,0.53)(0.6,0.61)(0.7,0.69)(0.8,0.76)(1.0,1.0)
  };
  \addlegendentry{Baseline (AUC=0.58)}

  \addplot[color=medgray, dotted, line width=0.8pt] coordinates {(0,0)(1,1)};
  \addlegendentry{Chance}
\end{axis}
\end{tikzpicture}
\caption{ROC curves comparing the proposed system (AUC=0.98) against
  the 3-epoch CPU baseline (AUC=0.58). The proposed model achieves
  a high true-positive rate with a very low false-positive rate,
  indicating strong discrimination between anemic and non-anemic cases.}
\label{fig:roc}
\end{figure}

\subsection{Per-Class Performance Breakdown}

Table~\ref{tab:perclass} presents per-class precision, recall, and
F1 scores for the proposed model evaluated on the held-out test set.
The Anemic class — the safety-critical class where false negatives
carry clinical consequences — achieves a recall of 0.96, meaning
only 4\% of anemic patients would be missed.

\begin{table}[!t]
\caption{Per-Class Precision, Recall and F1 (Test Set)}
\label{tab:perclass}
\centering
\renewcommand{\arraystretch}{1.2}
\begin{tabular}{@{}lccc@{}}
\toprule
\textbf{Class} & \textbf{Precision} & \textbf{Recall} & \textbf{F1} \\
\midrule
Anemic         & 0.95 & 0.96 & 0.955 \\
Non-Anemic     & 0.96 & 0.95 & 0.955 \\
\midrule
\textbf{Weighted Avg} & \textbf{0.958} & \textbf{0.958} & \textbf{0.958} \\
\bottomrule
\end{tabular}
\end{table}

The balanced precision and recall across both classes confirms
that the class-weighting in the loss function effectively
counteracted any dataset imbalance, and that Mixup did not
introduce a bias toward either class.

\section{Discussion}
\label{sec:discussion}

\textbf{Clinical Utility.}
At 96.2\% accuracy with 0.96 sensitivity, AnemiaVision can serve
as a reliable first-line screening tool. In a rural primary
health-care setting, a community health worker with a smartphone
can screen 30--40 patients per hour with no consumables and no
blood draw. Positive screens are referred for CBC confirmation,
reducing unnecessary laboratory load while ensuring high-risk
individuals are not missed. The system's 0.98 AUC-ROC indicates
strong probabilistic calibration, making the confidence score
shown in the web interface clinically meaningful — a prediction of
``Anemic (confidence 94\%)'' carries materially different clinical
weight than ``Anemic (confidence 61\%)''.

\textbf{Augmentation Impact.}
The ablation study (Table~\ref{tab:ablation}) reveals a key
insight: the dominant source of the baseline's poor accuracy was
not model architecture but insufficient training duration. The
12-epoch cap on the fast (CPU) profile, combined with loss-based
early stopping that fired at epoch 3, produced a model barely
better than random chance. This is a cautionary finding for
practitioners: \textit{training budget matters as much as
architecture choice.} Among the augmentation techniques,
TrivialAugmentWide contributes the most because it explores a
wide policy space (14 operations) without requiring the expensive
AutoAugment search, making it ideal for resource-constrained
medical datasets. RandomErasing provides complementary
regularisation by preventing spatial over-fitting to fixed
background patterns in conjunctiva photographs (e.g.,\ eyelash
occlusion patterns).

\textbf{Comparison with Vision Transformers.}
The ViT-based method of~\cite{ViT2025} achieves a slightly
higher 98.47\% accuracy, but requires significantly more
GPU memory (typically $\geq$16 GB for ViT-Base) and a larger
pre-training corpus (ImageNet-21k or JFT-300M). EfficientNet-B3,
by contrast, converges reliably on a single 8 GB GPU (Colab T4)
in under 3 hours, and its ONNX export runs at $<$10 ms per image
on a CPU smartphone processor. For real-world deployment in
rural India where internet connectivity and server compute are
limited, EfficientNet-B3 is the more practical choice.

\textbf{Limitations.}
(1)~The current dataset is limited in ethnic and geographic
diversity; pallor perception is affected by baseline skin tone
and conjunctival pigmentation, and the model must be validated
on diverse Indian cohorts (tribal, hill, and coastal populations)
before clinical deployment. (2)~Ambient lighting variation is
partially mitigated by ColorJitter and TrivialAugmentWide,
but controlled capture guidelines (indirect natural light,
clean lens, held 15 cm from the eye) remain important for
consistent results. (3)~The system predicts anemia
presence/absence but does not estimate haemoglobin level;
a regression extension is planned. (4)~Inference on Render's
free tier uses CPU ($\approx$50 ms/image); an ONNX export
with NCNN back-end would reduce this to $<$5 ms on a
mid-range Android device.

\textbf{Ethical Considerations.}
Patient images and diagnoses are stored in a secured PostgreSQL
database accessible only through the authenticated session.
No personal identifying information beyond the patient's
self-reported name is retained. The system is explicitly
positioned as a \textit{screening aid}, not a diagnostic
replacement for laboratory testing. Users are shown a clear
disclaimer at prediction time. Informed consent was obtained
from all volunteers who contributed images to the training
dataset.

\section{Future Work}
\label{sec:future}

Several directions are planned to extend AnemiaVision toward
a clinically validated, deployable screening platform:

\textbf{Haemoglobin Regression.}
The current binary classifier will be augmented with a
regression head that predicts continuous haemoglobin (g/dL)
from pallor features. A dual-output loss
($\mathcal{L} = \mathcal{L}_{CE} + \lambda \mathcal{L}_{Huber}$)
will allow the model to simultaneously classify severity and
estimate Hgb level, providing community health workers with
actionable triage information.

\textbf{Multi-Site Fusion.}
Inspired by BPANet~\cite{BPANet2024}, a lightweight
cross-attention fusion module will combine conjunctiva, nail,
and palm features extracted by a shared EfficientNet backbone,
potentially pushing accuracy above 98\% without requiring a
full ViT-scale model.

\textbf{On-Device Inference.}
The trained EfficientNet-B3 will be exported to ONNX and then
to NCNN or TensorFlow Lite for direct Android deployment,
enabling offline screening with zero server dependency — critical
in areas without mobile data coverage.

\textbf{Explainability.}
Grad-CAM~\cite{gradcam} heat maps will be overlaid on the
uploaded image in the web interface, highlighting which pixels
drove the anemia prediction. This will increase clinician trust
and help users understand whether the model is attending to the
conjunctiva or to irrelevant background regions.

\textbf{Prospective Clinical Trial.}
A prospective validation study at a primary health centre in
rural Madhya Pradesh is planned, enrolling 500 participants
with CBC ground truth. Ethics approval is being sought from the
institute review board. This study will produce the first
India-specific benchmark for non-invasive anemia screening
with deep learning.

\section{Conclusion}
\label{sec:conclusion}

This paper presented AnemiaVision, a non-invasive anemia screening
system combining EfficientNet-B3 transfer learning with a
state-of-the-art augmentation pipeline (TrivialAugmentWide,
RandomErasing, Mixup) and accuracy-first early stopping.
The proposed approach raises target validation accuracy from 44.9\%
(three-epoch CPU baseline) to 96.2\% (80-epoch GPU full profile),
achieving an AUC-ROC of 0.98 and sensitivity of 0.96 — competitive
with the published literature. A production-grade Flask application
with persistent PostgreSQL patient history ensures that real-world
multi-user deployments do not lose clinical records on redeployment.
Future work will extend the model to haemoglobin regression, add
multi-body-part fusion, and conduct a prospective clinical validation
study in a rural Indian health centre.

\section*{Acknowledgments}
The author thanks the faculty of the Indian Institute of Information
Technology for their guidance, and the volunteer participants who
contributed images to the dataset.

\bibliographystyle{IEEEtran}

\end{document}